\documentclass[runningheads]{llncs}
\usepackage{graphicx}
\usepackage{amsmath}
\usepackage{lipsum}
\usepackage{amssymb}
\usepackage{booktabs,multirow}
\usepackage{bibentry}
\usepackage{cite}
\usepackage[utf8]{inputenc}
\usepackage[symbol]{footmisc}
\begin{document}
\title{A Framework For Image Synthesis Using Supervised Contrastive Learning}
%
%
\author{Yibin Liu$^*$\and
Jianyu Zhang$^*$ \and
Li Zhang \and
Shijian Li$^\dagger$ \and
Gang Pan}

%
%
\institute{Zhejiang University\\
\email{\{yibinliu, jianyu.zhang, zhangli85, shijianli, gpan\}@zju.edu.cn}
}
%
\maketitle              
\def\thefootnote{*}\footnotetext{Equal Contribution}
\def\thefootnote{$\dagger$}\footnotetext{Corresponding Author}
\begin{abstract}
Text-to-image (T2I) generation aims at producing realistic images corresponding to text descriptions. Generative Adversarial Network (GAN) has proven to be successful in this task. Typical T2I GANs are 2-phase methods that first pre-train an inter-modal representation from aligned image-text pairs and then use GAN to train image generator on that basis. However, such representation ignores the inner-modal semantic correspondence, e.g. the images with same label. The semantic label in priory describes the inherent distribution pattern with underlying cross-image relationships, which is supplement to the text description for understanding the full characteristics of image. 
In this paper, we propose a framework leveraging both inter- and inner-modal correspondence by label guided supervised contrastive learning. We extend the T2I GANs to two parameter-sharing contrast branches in both pre-training and generation phases. This integration effectively clusters the semantically similar image-text pair representations, thereby fostering the generation of higher-quality images. 
We demonstrate our framework on four novel T2I GANs by both single-object dataset CUB and multi-object dataset COCO, achieving significant improvements in the Inception Score (IS) and Fréchet Inception Distance (FID) metrics of image generation evaluation. Notably, on more complex multi-object COCO, our framework improves FID by 30.1\%, 27.3\%, 16.2\% and 17.1\% for AttnGAN, DM-GAN, SSA-GAN and GALIP, respectively. We also validate our superiority by comparing with other label guided T2I GANs. The results affirm the effectiveness and competitiveness of our approach in advancing the state-of-the-art GAN for T2I generation.

\end{abstract}

\section{Introduction}
Text-to-image (T2I) generation targets on generating realistic images that match the corresponding text description. This captivating task has gained widespread attention and popularity owing to its vast creative potentials in art generation, image manipulation, virtual reality and computer-aided design.

T2I generation methods based on Generative Adversarial Network (GAN) \cite{DBLP:conf/nips/GoodfellowPMXWOCB14} have shown promising results. The typical approach can be decomposed the pre-training phase and GAN phase. They first pre-train the image and text features into a joint representation space, which provides effective understanding of the relationship between text descriptions and visual contents, and then use noval GAN to training the image generator on basis of joint representation. Since the introduction of notable AttnGAN \cite{xu2018attngan}, many subsequent works have utilized the Deep Attentional Multimodal Similarity Model (DAMSM) which employs contrastive learning to pull the paired image and text representations close while pushing away the unpaired ones. Consequently, DAMSM improve the consistency between image and text representations, resulting in effective downstream generation \cite{xu2018attngan, zhu2019dm, qiao2019mirrorgan, DBLP:conf/cvpr/LiaoHYR22}. Despite contrasting on the inter-modal text-image pair, each image sample may have specific category of similar samples that being ignored or pushed away, resulting in scrapping the underlying inner-modal distribution. Moreover, a brief textual description is usually insufficient to describe all the characteristics of an image. UniCL \cite{DBLP:conf/cvpr/YangLZXLYG22} proposes a unified contrastive loss in image-text-label space to leverage label information during representation learning. However, UniCL does not consider the rareness of samples with the same label in a batch, and is only applicable to single-label datasets.

Taking the inner-modal semantic into consideration, we introduce supervised contrastive learning into T2I GAN by referring to the categorical information of images, which enhances both the representation encoders and GAN generator, thereby improving the quality of image generation. For single-object image generation, we incorporate single-label supervised contrastive learning \cite{khosla2020supervised}. During the pre-training phase, our proposed supervised contrastive loss leverages additional image labels to group the representations for image and text of the same class while distinguishing images of different classes. During the GAN phase, we also employ the supervised contrastive loss to simultaneously increase the synthetic images' similarities of same class and the matching degree to their text pair. For multi-object image generation, we leverage same approach on single-object scenario by changing the supervised contrastive loss to multi-label case \cite{malkinski2022multi}. We evaluate our method on datasets CUB \cite{2011The} and COCO \cite{lin2014microsoft}. By comparing to four base models: AttnGAN \cite{xu2018attngan}, DM-GAN \cite{zhu2019dm} SSA-GAN \cite{DBLP:conf/cvpr/LiaoHYR22} and GALIP \cite{DBLP:conf/cvpr/TaoB0X23}, our experiments show that our method is capable of improving the quality of generated images measured by common metrics: the Inception Score (IS) \cite{salimans2016improved} and Fréchet Inception Distance (FID) \cite{unterthiner2017coulomb}.

The contributions of our work can be summarized as follows:
\begin{itemize}
\item We incorporate supervised contrastive learning to T2I generation which encourages the inherent data distribution patterns delineated by semantic labels, thereby enhancing the generation of coherent and faithful images. 
\item Our framework employs two symmetric parameter-sharing branches in the pre-training and GAN phase of T2I generation, which is compatible for single- and multi-object contrastive learning by corresponding loss. Such extension converges image representations carrying same semantics within proximity in the pre-training phase, which enables the GAN generator to glean insights from a broader spectrum of related data instances. 
\item Our framework can improve famous T2I GANs' generation quality on both single-object CUB and multi-object COCO dataset. Most notably, on more complex COCO dataset, our framework improves the FID of AttnGAN, DM-GAN, SSA-GAN and GALIP by 30.1\%, 27.3\%, 16.2\% and 17.1\% , respectively. We also demonstrate the superiority of our framework comparing with other label guidance options.
\end{itemize}

\section{Related Work}

\subsection{Contrastive Learning}
Contrastive learning is a self-supervised method which has been successful in representation learning. It plays a crucial role in serving computer vision tasks and extends influence to other research field like natural language processing. Contrastive learning follows the intuition that similar data samples should be closer in the representation space, while dissimilar samples should be far apart. 
Typical contrastive learning setting SimCLR \cite{chen2020simple} augments image into two randomly warped views and extracts their representations through twin encoders. The two branches of representation are then projected to same feature space to apply contrastive loss \cite{DBLP:journals/corr/abs-1807-03748}, where the paired view of image is considered as positive sample and vice verca. Other variants of contrastive learning mainly differ in the formulation of negative samples \cite{DBLP:conf/cvpr/He0WXG20}, the asymmetric design of twin encoders\cite{DBLP:conf/nips/GrillSATRBDPGAP20}, or contrastive loss definition\cite{DBLP:conf/icml/ZbontarJMLD21}. All these methods have either comparable results or exceed supervised methods on many representation learning benchmarks \cite{DBLP:conf/cvpr/DengDSLL009}. In addition to construct the positive and negative samples by self supervision, researchers \cite{khosla2020supervised,malkinski2022multi} also utilize image classification labels to formulate single- and multi-label contrastive loss, the former achieves high accuracy in image classification while the latter succeeds in visual reasoning. 
Contrastive learning has also been explored to bridge the modality gap and create unified representation for multi-modal pre-training. Trained by fine-curated large scale image text pairs, CLIP \cite{DBLP:conf/icml/RadfordKHRGASAM21} has demonstrated great zero-shot capability for dozens of visual and image-text downstream tasks. 

These contrastive learning progresses proves the feasibility of aligning different feature views at low annotation cost. We adopt the intuition that any data representation can be improved by referencing similar semantic concepts from both inter- and inner-modal data, therefore our framework designs multiple ways of feature alignment which will be detailed in Section \ref{sec:method}.

\subsection{GAN for Text-to-Image Generation} \label{sec:T2IGAN}
In recent years, image generation has experienced rapid development starting from the remarkable success of Generative Adversarial Network (GAN) which trains a generative model by adversarial discrimination \cite{zhang2017stackgan, zhang2018stackgan++, xu2018attngan, zhu2019dm, qiao2019mirrorgan, DBLP:conf/cvpr/Tao00JBX22, DBLP:conf/cvpr/LiaoHYR22}. Reed et al. \cite{reed2016generative} were the first to employ GAN to generate images from text descriptions. To synthesize higher resolution images, Zhang et al. propose the StackGAN \cite{zhang2017stackgan} and StackGAN++ \cite{zhang2018stackgan++} employing a multi-generator strategy that first generates a low-resolution image and then finetunes followup generators to produce high resolution realistic images. Many works follow this multi-stage stack structure \cite{xu2018attngan, zhu2019dm, qiao2019mirrorgan,yin2019semantics, ruan2021dae} to improve image generation quality. On basis of StackGAN++, AttnGAN \cite{xu2018attngan} introduced attention mechanism to refine the process of generating images from fine-grained textual descriptions at different stages of image generation. In addition, AttnGAN proposed the Deep Attentional Multi-modal Similarity Model (DAMSM) to improve multi-granular consistency between image and text. DM-GAN \cite{zhu2019dm} proposed dynamic memory to store the intermediate generated images and retrieve the most relevant textual information with gated attention to update the image representation accordingly. 

Although the multi-stage GAN is designate for high-resolution progressive image generation, its training complexity grows as the stage stacking. To overcome this, DF-GAN \cite{DBLP:conf/cvpr/Tao00JBX22} proposed single-stage generation, whose generator uses a series of UPBlock specially designed for high resolution feature upsampling. DF-GAN further used Matching-Aware Gradient Penalty and hinge loss to train the UPBlocks. Followup SSA-GAN \cite{DBLP:conf/cvpr/LiaoHYR22} used a Semantic Spatial Aware Convolution Network (SSACN) block to predict text aware mask maps based on the current generated image features, which facilitates the fusion and consistency between image and text. These conventionally designed single-stage methods greatly reduce the complexity of T2I generation, meanwhile others seek for utilizing famous visual-language pre-training 
techniques to bridge the inter-modal gap. GALIP \cite{DBLP:conf/cvpr/TaoB0X23} directly integrates CLIP \cite{DBLP:conf/icml/RadfordKHRGASAM21} to harness the well-aligned image-text representation and extend GAN's ability to synthesize complex images. Hui et al. \cite{ye2021improving} propose a framework leveraging contrastive learning to enhance the consistency between caption generated images and the originals. 
All these T2I GANs focus on the inter-modal image text alignment without considering inner-modal association, which in some extent leads to flaws in the generation results. Our framework instead encourages both inter- and inner-modal association.


\begin{figure}[t]
    \centering
    \includegraphics[width=0.8\textwidth]{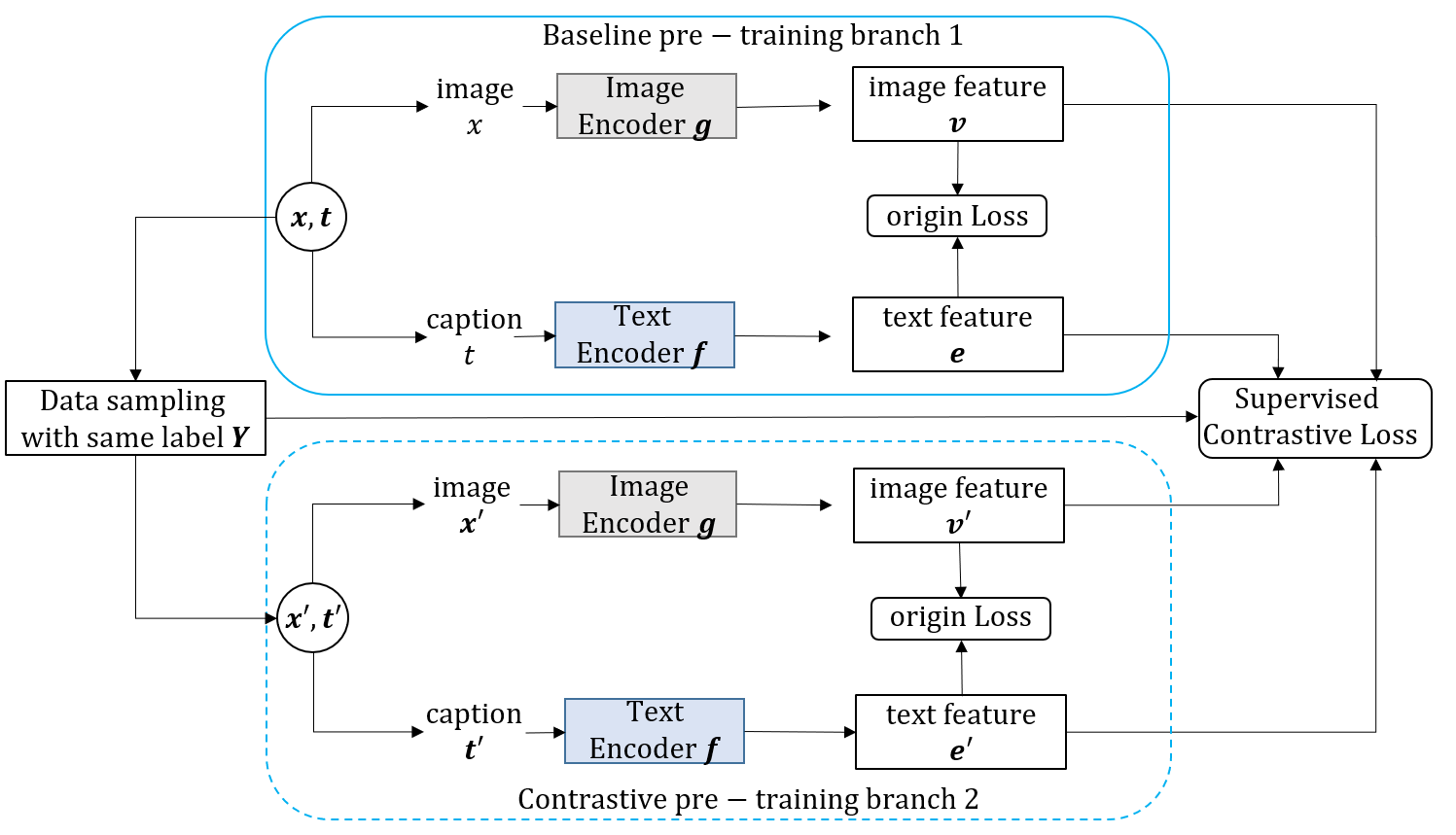}
    \caption{Pre-training phase. Our data sampling strategy initiates two contrast branches with shared parameters to separately encode the image-text pairs of same label. The original Loss is consistent to the method our framework applied on. The supervised contrastive loss works on quadruple of image and text representations from both branches.}
    \label{fig1}
\end{figure}

\section{Method} \label{sec:method}
In this section, we introduce a simple effective framework which integrates supervised contrastive learning to leverage the inner-modal association, thereby enhancing the generation quality of T2I GANs. Like novel contrastive learning approach, we adopt the dual tower structure and create two symmetric branches of contrast opponents for both pre-training and GAN phases. In pre-training phase, the supervised contrastive learning encourages the representation coherency for image-text pairs sharing same semantics. In favor of the coherent representation, in the GAN phase, the supervised contrastive learning establishes additional guidance for the semantic consistency of the generated images. We detail our framework adaptation and enhanced T2I GAN learning objectives for the two phases in the following respective sections.

\subsection{Supervised Contrastive Learning for Pre-training}
Typical T2I GANs pre-train the image and text encoders by maximizing the paired image-text representation similarity and the unpaired dissimilarity. To enhance this learning process, we extend the pre-training by supervised contrastive learning on the image-text pair with shared label. 
The extension has three components shown in Figure \ref{fig1}.

\subsubsection{Data Sampling Strategy}
At each training step, we randomly sample a batch of $N$ examples which consist of $N$ captions $\boldsymbol{t}$ , corresponding images $\boldsymbol{x}$ and label set $\boldsymbol{Y}$. To construct contrastive pair, we ensure that each sample has reference example with the same labels: for each sample ($t_i$, $x_i$, $Y_i$), we select a sample ($t_{i}'$, $x_{i}'$, $Y_{i}'$) as its pair where $Y_i \cap Y_{i}' \neq \emptyset$. 

\subsubsection{Image Encoder $\boldsymbol{g}$ And Text Encoder $\boldsymbol{f}$}
In pre-training phase, the encoder extracted representations usually have multi-granular features to encourage the deep fusion, e.g., the global/local views of image, and the sentence/word level of text. Our methods do not change the functionalities but extend them by applying shared image and text encoders $\boldsymbol{g, f}$ to extract contrastive pair image representations $\boldsymbol{v = g(x)}, \boldsymbol{v' = g(x')}$ and text representations $\boldsymbol{e=f(t)}, \boldsymbol{e'=f(t')}$. Our framework is indifferent for the type of encoders, where we keep them consistent to the baseline methods our framework applied to. Specifically, for AttnGAN \cite{xu2018attngan}, DM-GAN \cite{zhu2019dm} and SSA-GAN \cite{DBLP:conf/cvpr/LiaoHYR22}, we use Inception-v3 \cite{DBLP:conf/cvpr/SzegedyVISW16} as image encoder $\boldsymbol{g}$ and Bi-LSTM \cite{DBLP:journals/tsp/SchusterP97} as text encoder $\boldsymbol{f}$. For GALIP \cite{DBLP:conf/cvpr/TaoB0X23}, we use transformer-based CLIP image and text encoders. The weights of the text encoder and image encoder are frozen during the training phase of the GAN.

\subsubsection{Learning Objective}
With the data sampling strategy, we define the objective for training. For image-text matching using Inception-v3 and Bi-LSTM, we consider ($t_i$, $x_i$) and ($t_{i}'$, $x_{i}'$) as positive image-text pairs to calculate DAMSM loss same as AttnGAN \cite{xu2018attngan}. As for CLIP encoder, we use symmetric cross entropy loss \cite{DBLP:conf/icml/RadfordKHRGASAM21}. To apply supervised contrastive loss, we formulate positive pairs from sampling strategy for image-image, image-text and text-text associations. Specifically, ($t_i$, $t_{i}'$), ($t_i$, $t_j$) and ($t_i$, $t_{j}'$) are considered as positive text-text pairs where $Y_i \cap Y_j \neq \emptyset$. It is worth noting that in single-object dataset CUB, each corresponding image-text sample only has one label, while in complex multi-object dataset COCO, it has multiple labels. Therefore, we use different supervised contrastive loss functions to deal with different label sharing.

For \textbf{one label} scenario, we treat sample pairs with the same label as positive pairs and apply single-label supervised contrastive loss. Given a random batch of $N$ instances, we pick $2N$ instances after data sampling stategy where each instance is guaranteed to have at least one same label in other instances. In order to facilitate the calculation, we concatenate the sampled instances with the original ones to obtain the image representation $\boldsymbol{\widetilde{v}} =\{\boldsymbol{v},\boldsymbol{v'}\}$, text representation $\boldsymbol{\widetilde{e}} =\{\boldsymbol{e},\boldsymbol{e'}\}$ and labels $\boldsymbol{\widetilde{Y}} =\{\boldsymbol{Y},\boldsymbol{Y'}\}$ at this step. Let $sim(a,b) = a^{T}b/(||a||\cdot||b||)$ denote the cosine similarity between $a$ and $b$. For a certain representation $u_i$ and its relative batch of representations $\boldsymbol{w}$, the supervised contrastive loss function is calculated as
\begin{equation}
    \mathcal{L}^{sup}(u_i, \boldsymbol{w}) = \frac{-1}{|P_{s}(i)|}\sum_{p\in P_s(i)} log \frac{exp(sim(u_i, w_p)/\tau)}{\sum_{j \neq i}^{2N} exp(sim(u_i, w_j)/\tau)}
\end{equation}
where $P_{s}(i)=\{p \in \{1,...,2N\}:\widetilde{Y_p}=\widetilde{Y_i} \}$ is the set of indices of all positives in the batch distinct from $i$, $|P_{s}(i)|$ is the cardinality of $P_{s}(i)$ and $\tau$ denotes the temperature parameter. We can specifically compute supervised contrastive losses for image-image $\mathcal{L}^{sup}_{img}$, text-text $\mathcal{L}^{sup}_{txt}$ and image-text $\mathcal{L}^{sup}_{i2t}$ as follows:
\begin{equation}
    \mathcal{L}^{sup}_{img} = \sum_{i=1}^{2N}\mathcal{L}^{sup}(\widetilde{v_i}, \boldsymbol{\widetilde{v}})
\end{equation}
\begin{equation}
    \mathcal{L}^{sup}_{txt} = \sum_{i=1}^{2N}\mathcal{L}^{sup}(\widetilde{e_i}, \boldsymbol{\widetilde{e}})
\end{equation}
\begin{equation}
    \mathcal{L}^{sup}_{i2t} = \sum_{i=1}^{2N}\mathcal{L}^{sup}(\widetilde{e_i}, \boldsymbol{\widetilde{v}}) + \sum_{i=1}^{2N}\mathcal{L}^{sup}(\widetilde{v_i}, \boldsymbol{\widetilde{e}})
\end{equation}

Similarly, for \textbf{multi-label} scenarios, we consider instances that have one or more common labels as positive pair. We employ multi-label supervised contrastive loss, which replaces $P_{s}(i)$ with $P_{m}(i)=\{p \in \{1,...,2N\}:\widetilde{Y_p} \cap \widetilde{Y_i}\neq \emptyset\}$ in the calculation process while keeping all other calculation the same as in the single-label contrastive loss.

The final objective function for the pre-training phase is a co-op of origin loss and supervised contrastive loss
\begin{equation}
    \mathcal{L}_{pre} = \mathcal{L}_{orgin} + \lambda_1(\mathcal{L}^{sup}_{img} +\mathcal{L}^{sup}_{txt}+\mathcal{L}^{sup}_{i2t} )
\end{equation}
where $\lambda_1$ is the weight of supervised contrastive loss. Depending on the baseline GAN methods, $\mathcal{L}_{orgin}$ can either be DAMSM or symmetric cross entropy loss.

\begin{figure*}[t]
    \centering
    \includegraphics[width=0.9\textwidth]{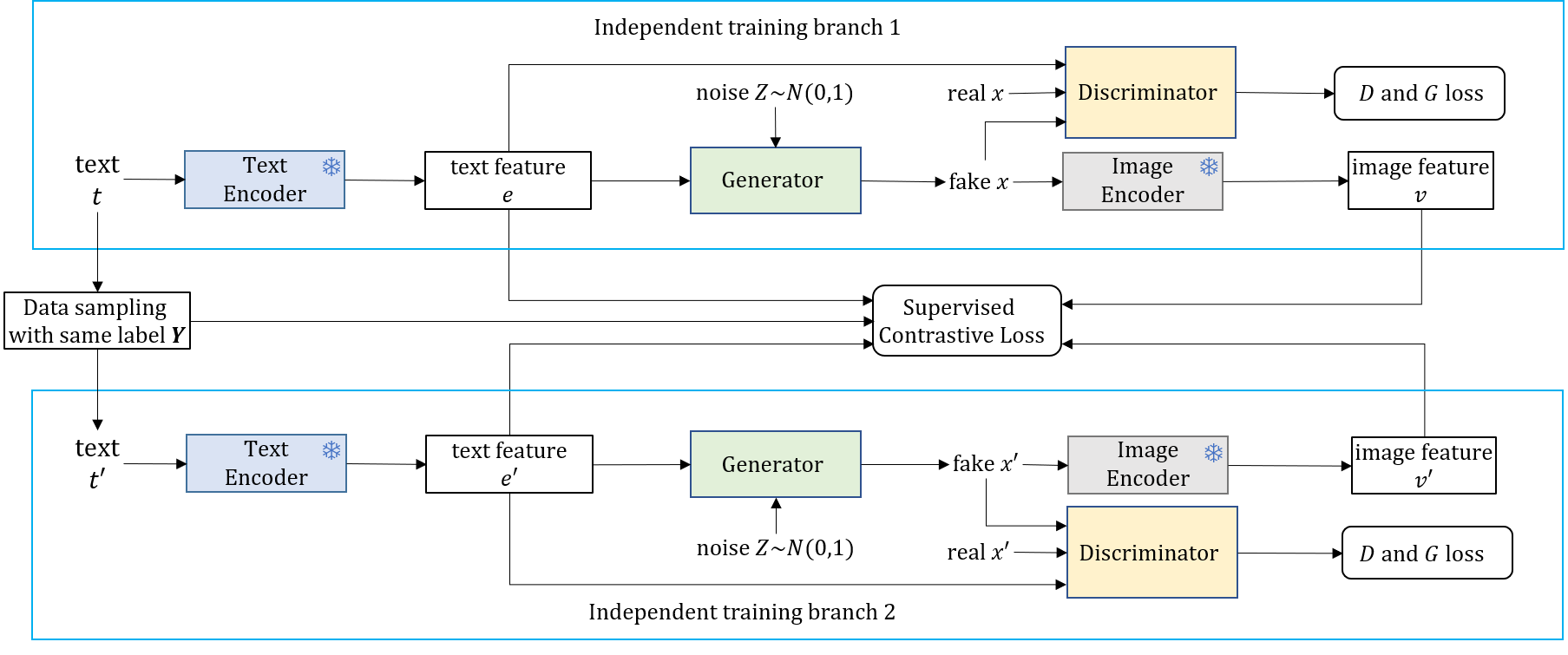}
    \caption{GAN training phase. Same as pre-training phase, we use two parameter-sharing T2I GAN branches to contrast the text-image pairs sharing same label. The supervised contrastive loss is performed on quadruple of text and generated fake image representations from two branches. In this phase, the pre-trained encoders are inference-only. }
    \label{fig2}
\end{figure*}

\subsection{Supervised Contrastive Learning for GAN}
Intuitively, shared labels reflect common visual semantics within the images. In captioning datasets, the brief text annotation typically use concise descriptions to depict partial aspect of images. Therefore, during generator training, we provide instances sharing same label to encourage the generator to refer to the similar instances. Our generator training framework is illustrated in figure \ref{fig2}.

\subsubsection{Data Sampling Strategy}
Same as the pre-training phase, we sample a batch of images $\boldsymbol{x}$ and $\boldsymbol{x'}$, text captions $\boldsymbol{t}$ and $\boldsymbol{t'}$, labels $\boldsymbol{Y}$ and $\boldsymbol{Y'}$. The captions are extracted to text representations $\boldsymbol{e}$ and $\boldsymbol{e'}$ by pre-trained text encoder $\boldsymbol{f}$.

\subsubsection{GAN Adaptation} 
As discussed in Section \ref{sec:T2IGAN}, the mainstream T2I GAN methods are based on two types: the multi-stage StackGAN series \cite{zhang2018stackgan++} and the one-stage DFGAN \cite{DBLP:conf/cvpr/Tao00JBX22}. Our framework can be applicable to both types. Given the ground-truth real image $\boldsymbol{x}$, the generator $G$ utilizes text representations $(\boldsymbol{e},\boldsymbol{e'})$ and noise $z$ to generate fake images $(\boldsymbol{x_f},\boldsymbol{x_f'})$ in two branches. Subsequently, the discriminator calculates the generator losses $(\mathcal{L}_G^o, \mathcal{L}_G^{o'})$ and discriminator losses $(\mathcal{L}_D^o, \mathcal{L}_D^{o'})$ for two branches from $(\boldsymbol{x}, \boldsymbol{e}, \boldsymbol{x_f})$ and $(\boldsymbol{x'}, \boldsymbol{e'}, \boldsymbol{x_f'})$, respectively. Meanwhile, the generated images from both branches are encoded by an image encoder and obtains fake image representations $(\boldsymbol{v_f}, \boldsymbol{v_f'})$. These representations are then paired with $(\boldsymbol{e}, \boldsymbol{e'})$ to calculate supervised contrastive loss.

\subsubsection{Learning Objective}
In our framework, the objective function for discriminator loss during the training process is identical to the GAN baselines in both branches, and the overall discriminator loss $\mathcal{L}_D$ is the sum of loss from two branches. As for the generator loss $\mathcal{L}_G$, one-stage GAN typically use conditional generation loss \cite{DBLP:conf/cvpr/Tao00JBX22,DBLP:journals/corr/LimY17} while multi-stage GAN often incorporate additional non-conditional generation loss \cite{zhang2018stackgan++}. Our method does not vary the usage of baseline generator losses but adding extra supervised contrastive losses for image-to-image and image-text pairs.

Similar to pre-training phase, for sampled batch, we first concatenate the generated fake image representation $\boldsymbol{\overline{v}} =\{\boldsymbol{v_f},\boldsymbol{v_f'}\}$, the corresponding text representations $\boldsymbol{\widetilde{e}} =\{\boldsymbol{e},\boldsymbol{e'}\}$ and the labels $\boldsymbol{\widetilde{Y}} =\{\boldsymbol{Y},\boldsymbol{Y'}\}$. 
The discriminator and generator loss function are then computed as follows:

\begin{equation}
    \mathcal{L}_D = \mathcal{L}_{D}^{o} + \mathcal{L}_{D}^{o'}
\end{equation}
\begin{equation}
    \mathcal{L}_G = \mathcal{L}_{G}^{o} + \mathcal{L}_{G}^{o'} + \lambda_2(\mathcal{L}^{sup}_{img}+\mathcal{L}^{sup}_{i2t} )
\end{equation}
where
\begin{equation}
    \mathcal{L}^{sup}_{img} = \sum_{i=1}^{2N}\mathcal{L}^{sup}(\overline{v_i}, \boldsymbol{\overline{v}})
\end{equation}
\begin{equation}
    \mathcal{L}^{sup}_{i2t} = \sum_{i=1}^{2N}\mathcal{L}^{sup}(\widetilde{e_i}, \boldsymbol{\overline{v}}) + \sum_{i=1}^{2N}\mathcal{L}^{sup}(\overline{v_i}, \boldsymbol{\widetilde{e}})
\end{equation} and $\lambda_2$ is the weight of supervised contrastive loss.

\section{Experiments}
We choose novel multi-stage (AttnGAN, DM-GAN) and one-stage (SSA-GAN, GALIP) GANs to validate the superiority and universality of our framework on T2I generation for both single-object CUB \cite{2011The} and multi-object COCO \cite{lin2014microsoft} datasets. We also conduct extensive ablations to assess the effectiveness of each component our framework proposes.


\subsubsection{Evaluation Metric}
We follow the baselines' evaluation protocol on the CUB and COCO datasets, which uses Inception Score (IS) \cite{salimans2016improved} and Fréchet Inception Distance (FID) \cite{unterthiner2017coulomb} as quantitative evaluation metrics. 
After training completion, we generate 30,000 images in resolution 256×256 on the test set and compute IS and FID scores. Several previous works \cite{DBLP:conf/cvpr/LiZZHHLG19, DBLP:conf/cvpr/Tao00JBX22} have pointed out that IS can not provide useful guidance to evaluate the quality of the synthetic images on dataset COCO, thus we only evaluate IS on CUB dataset. Since GALIP was not evaluated on IS, we only compared with GALIP on FID.

\subsubsection{Implementation Details}
We apply our framework to four novel baselines (AttnGAN, DM-GAN, SSA-GAN and GALIP) on both CUB and COCO datasets. During pre-training phase, we set $\lambda_1$ to 0.5 for CUB and 0.05 for COCO. For GAN phase, we set $\lambda_2$ of the four baselines to 5, 2.5, 0.2, 0.15 for CUB and 2.5, 2.5, 0.1, 0.15 for COCO. The training epochs of the four baselines are 600, 800, 600, 2000 for CUB and 120, 200, 120, 2000 for COCO. Our training uses 1, 1, 3, 3 NVIDIA GeForce RTX 3090 GPU respectively for the four baselines.

\begin{figure*}[t]
    \centering
    \includegraphics[width=1\textwidth]{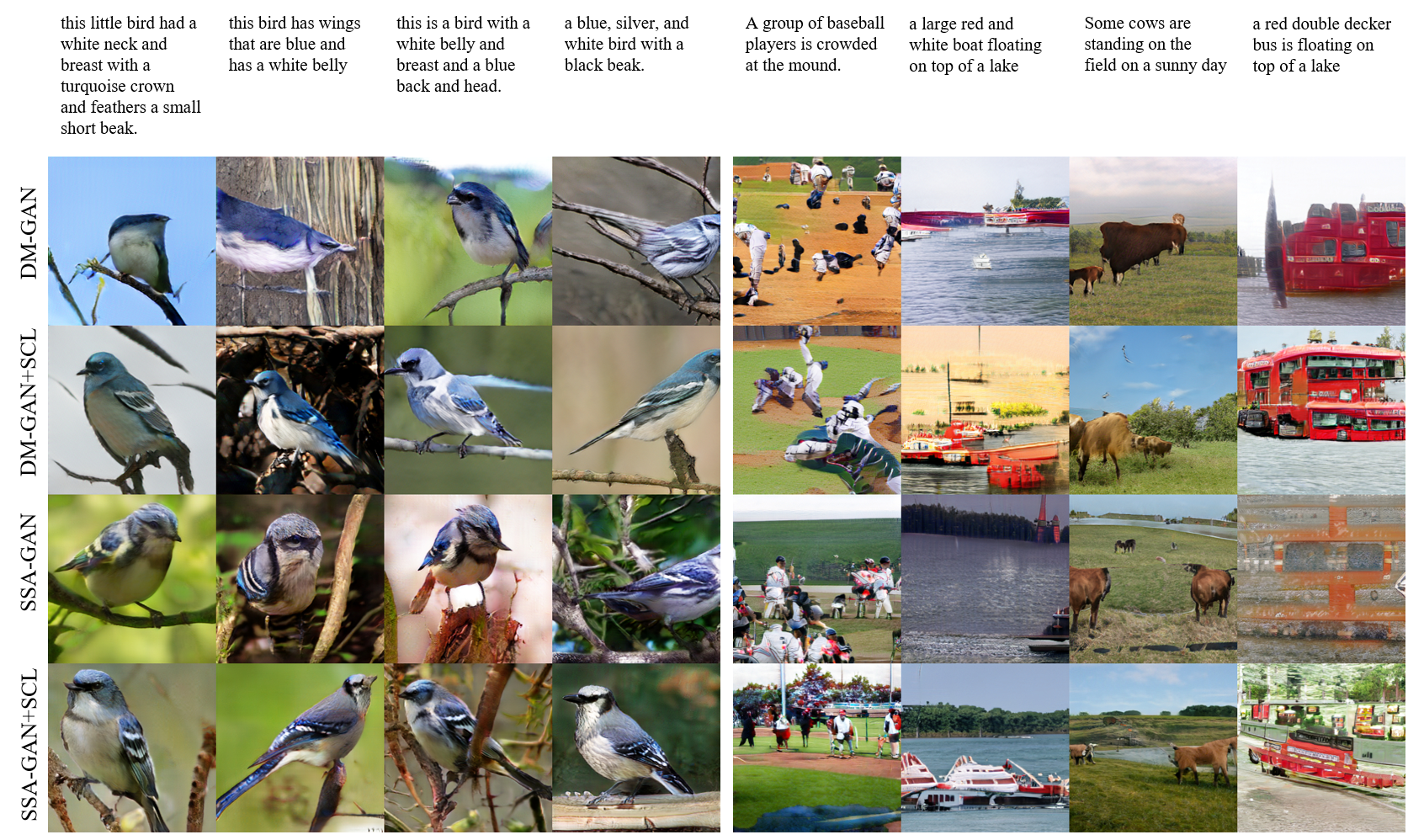}
    \caption{Qualitative comparison on CUB and COCO datasets for DM-GAN and SSA-GAN baselines w/o the utilization of our framework (denoted as "+SCL"). The input text descriptions are given in the first row and the corresponding generated images from different methods are shown in the same column. The left 4 columns are from CUB, and right 4 columns from COCO.}
    \label{fig3}
\end{figure*}

\begin{table}[!htbp] 
	\caption{Performance of IS and FID of AttnGAN, DM-GAN, SSA-GAN and
these models with our framework increment on the CUB and COCO test set. ↑ denotes higher values indicate better quality. ↓ denotes lower values indicate better quality. * denotes results obtained from publicly released pre-trained models by the authors. "+SCL" represents the model trained by our framework. \textbf{Bold} for better performance.}
	\centering
	\label{table1}
	\begin{tabular}{cccc}
		\toprule
		\multirow{2}{*}{\centering Methods} & \multicolumn{2}{c}{CUB} & \multicolumn{1}{c}{COCO} \\
		\cmidrule(lr){2-3}\cmidrule(lr){4-4}
            & IS↑ & FID↓ & FID↓ \\
		\midrule
		AttnGAN*   & 4.36±.03 & 23.98 & 33.10 \\
		AttnGAN+SCL & \textbf{4.61±.06} & \textbf{17.83} & \textbf{23.14} \\
            \midrule
		DM-GAN*   & 4.65±.05 & 15.31 & 26.56 \\
		DM-GAN+SCL & \textbf{4.95±.05} & \textbf{14.35} & \textbf{19.32} \\
            \midrule
		SSA-GAN*   & 5.07±.08 & 15.69 & 19.37 \\
		SSA-GAN+SCL & \textbf{5.14±.09} & \textbf{14.20} & \textbf{16.24} \\
            \midrule
		GALIP   & - & 10.08 & 5.85 \\
		GALIP+SCL & - & \textbf{9.90} & \textbf{4.85} \\
		\bottomrule
	\end{tabular}
\end{table}

\subsection{Quantitative Results}
The four baselines and our enhancement results are reported in Table \ref{table1}.
On single-object CUB dataset, our framework is able to improve the IS of AttnGAN by 5.7\%, DM-GAN by 6.5\%, and SSA-GAN by 1.4\%. These results demonstrate that our framework effectively improves the clarity and diversity of generated images. Moreover, our framework improves the FID of AttnGAN by 25.6\%, DM-GAN by 6.3\%, SSA-GAN by 9.5\%  and GALIP by 1.8\%. 
On more challenging multi-object COCO dataset, our framework is able to significantly improve the FID of all baselines. Specifically, we improves AttnGAN, DM-GAN, SSA-GAN and GALIP by 30.1\%, 27.3\%, 16.2\% and 17.1\% respectively. These results indicate that semantic relationship modeling is crucial for enhancing the T2I GAN generation quality, and the more complex scenario benefits more from it. 

\subsection{Visual Quality}
In this section, we further compare the visual quality of generated images by a subset of CUB and COCO datasets for DM-GAN, SSA-GAN baselines before and after applying our framework, which are shown in Figure \ref{fig3}.

For the CUB dataset, we randomly select text-generated images belonging to the "Tree Swallow" category for comparison. In the first and second column, the images generated by DM-GAN exhibit severe error in producing bird head, while DM-GAN with supervised contrastive learning generates natural bird images. SSA-GAN on the other hand can generate natural bird images, but the generated bird images do not always match the descriptions or the desired bird species. For example, the bird generated in the 1st column exhibits yellow and green wings, and the bird in the 3rd column had red tails, which are not mentioned in the text description and do not align with the characteristics of Tree Swallows. On the contrary, SSA-GAN enhanced by our framework can produce birds that match the text description specifying blue-black-white wings, and is consistent with the features of Tree Swallows. In addition, the images generated by our framework exhibit strong similarity for same species, which further confirms the validity of supervised contrastive learning.

Generating realistic and textually coherent images that align with the descriptions is more challenging in the COCO dataset. However, our framework outperforms the baseline in terms of generating higher quality and more textually consistent images. For example, in 6th column, both DM-GAN and SSA-GAN failed to generate a red boat mentioned in the input text, but DM-GAN and SSA-GAN enhanced by our framework successfully generate the desired object. In 8th column, the bus generated by SSA-GAN is orange-yellow which deviates from the "red" description, while SSA-GAN enhanced by our framework successfully produce a red bus matching the description.

\subsection{Ablation Study}
In both pre-training and GAN phases we incorporate image-image supervised contrastive loss $L^{sup}_{img}$ and image-text $L^{sup}_{i2t}$ supervised contrastive loss. In this section, we verify the effectiveness of \textit{pre}, $L^{sup}_{img}$ and $L^{sup}_{i2t}$ in our framework by conducting extensive ablation study on the CUB and COCO dataset in Table \ref{table2}.

\begin{table}[!htbp]
	\caption{Ablations of AttnGAN baseline. Our pre-trained encoders (\textit{pre}), image-image supervised contrastive loss ($L^{sup}_{img}$) and image-caption supervised contrastive loss ($L^{sup}_{i2t}$) are ablated independently. }
	\centering
	\label{table2}
	\begin{tabular}{ccccccc}
		\toprule
		\multirow{2}{*}{\centering ID} & \multicolumn{3}{c}{Components}& \multicolumn{2}{c}{CUB} & \multicolumn{1}{c}{COCO} \\
		\cmidrule(lr){2-4}\cmidrule(lr){5-6}\cmidrule(lr){7-7}
            &\centering $pre$&\centering $L^{sup}_{img}$&\centering $L^{sup}_{i2t}$& IS↑ & FID↓ & FID↓ \\
		\midrule
		1   &-&-&-& 4.36±.03 & 23.98 & 33.10 \\
		2 &\checkmark&-&-&  4.41±.05& 20.83 &  26.90\\
		3   &\checkmark&\checkmark&-&  4.53±.04& \textbf{17.42} & 24.14 \\
		4 &\checkmark&-&\checkmark&  4.45±.07& 18.53 & 25.09 \\
		5   &\checkmark&\checkmark&\checkmark&\textbf{4.61±.06} & 17.83 & \textbf{23.14} \\
		\bottomrule
	\end{tabular}
\end{table}

We consider the AttnGAN as the baseline (ID 1). When using pre-trained encoders (ID 2), all metrics get improved, which indicates that the encoders with supervised contrastive learning obtain image and text representations with better semantic alignment and consistency (the visualization of representation is given in supplementary material). Building upon \textit{pre}, introducing $L^{sup}_{img}$ (ID 3) and $L^{sup}_{i2t}$ (ID 4) individually also results in improvement for all metrics, which suggests that using $L^{sup}_{img}$ and $L^{sup}_{i2t}$ separately enhances the similarity between image-image and image-text representations with the same label. The usage of $L^{sup}_{img}$ shows better improvement comparing to $L^{sup}_{i2t}$, indicating that previous work is more lack of the intrinsic image modeling on dataset semantic level. However, when $L^{sup}_{img}$ and $L^{sup}_{i2t}$ are used together (ID 5), both IS of CUB and FID of COCO are improved, but the FID of CUB inferior a little. The reason is that $L^{sup}_{i2t}$ surges impact on facilitating text-image fusion and representation similarity, resulting in the IS improvement. On the other hand, when the encoded text features become more adaptive to the image features with same labels, the diversity of generated images also increases(more deeply constrained by the text descriptions with same label). Consequently, the FID slightly drops as it measures the KL divergence between the real images and generated images \cite{DBLP:conf/cvpr/LiaoHYR22}.

\subsection{Comparison to other label-supervised methods}

\begin{table}[!htbp] 
	\caption{AttnGAN baseline comparison of other semantic label integration options including UniCL, cross-entropy and ours. }
	\centering
	\label{table3}
	\begin{tabular}{cccc}
		\toprule
		\multirow{2}{*}{\centering Methods} & \multicolumn{2}{c}{CUB} & \multicolumn{1}{c}{COCO} \\
		\cmidrule(lr){2-3}\cmidrule(lr){4-4}
            & IS↑ & FID↓ & FID↓ \\
		\midrule
		AttnGAN*   & 4.36±.03 & 23.98 & 33.10 \\
            UniCL   & 4.39±.02 & 19.42 & 28.67 \\
            cross-entropy & 4.34±.05 & 21.15 & 27.30 \\
		Ours & \textbf{4.61±.06} & \textbf{17.83} & \textbf{23.14} \\
		\bottomrule
	\end{tabular}
\end{table}

To our best knowledge, there is no existing approach in this field leveraging labels information as additional guidance like our framework does. To demonstrate the novelty of our approach, we use two simple settings that commonly used for plug-in label learning as extra baselines. Firstly, we apply UniCL \cite{DBLP:conf/cvpr/YangLZXLYG22} to AttnGAN. On CUB dataset, UniCL can easily be adopted because each image only asscociates with one label. In order to apply UniCL to the COCO dataset, we replaced its single-label supervised contrastive loss to a multi-label supervised contrastive loss. Secondly, we introduce cross-entropy loss in classification task to AttnGAN. We introduce a pre-trained fully connected network as a image classifier and add the cross-entropy loss to the existing loss and train by multi-task learning. The results are shown in the table \ref{table3}. As the UniCL and cross-entropy improving the AttnGAN slightly, our framework demonstrate largest margin of visual enhancement for all metrics, indicating the compatibility of our framework with T2I GAN baselines.


\section{Conclusions}
In this work, we introduce a novel framework that harness semantic information with supervised contrastive learning to improve T2I GAN. Our framework use the two branch contrast to extend the original method across the pre-training and GAN phases. In pre-training phase, we employ label guided data sampling strategy, where we define positive pair as the images with same label. Driven by supervised contrastive loss on the positive image pairs and their corresponding text, the pre-training encoder elevates the representation similarity of images with same semantic concepts and push away those without. In the GAN phase, we first proceed original GAN for each branch independently and formulate a quadruple including the representations of generated positive image pair and their corresponding texts from two branches. We then employ augmented supervised contrastive loss to the quadruple which, like in pre-training phase, serves to elevate the similarity between images characterized by common semantic, thereby enhancing the image generation quality.

We apply our framework to famous four GAN baselines including AttnGAN, DM-GAN, SSA-GAN and GALIP and conduct experiments on single-object CUB and multi-object COCO dataset. The results demonstrate that our framework can indifferently improve baselines on both datasets with considerable margin, especially the more complex COCO.  

Although we only demonstrate the effectiveness on the datasets with detailed label annotation, our framework can be extended to other image-text pair only datasets by noun extraction from all text as labels, which will be the next step of our research interest. Recently, the advent of data-centric methodologies such as SAM \cite{kirillov2023segment} has further curtailed the expenses for semantic label acquisition, subsequently relaxing the prerequisites for implementing our framework. Furthermore, we expect this work to exhibit potential application for diffusion models especially on efficiency improving due to the adaptable nature of our framework. We defer the extension to future research endeavors.

\section{Acknowledgments}
This research was supported by STI 2030—Major Projects 2021ZD0200403. The authors like to thank the authors of DM-GAN for providing the details of its implemention and the anonymous reviewers for their review and comments.

\end{document}